\documentclass{article}
\usepackage{ijcai25}

\usepackage{times}
\usepackage{soul}
\usepackage{url}
\usepackage[hidelinks]{hyperref}
\usepackage[utf8]{inputenc}
\usepackage[small]{caption}
\usepackage{graphicx}
\usepackage{amsmath}
\usepackage{amsthm}
\usepackage{booktabs}
\usepackage{algorithm}
\usepackage{algorithmic}
\usepackage[switch]{lineno}

\usepackage{multirow}
\usepackage[normalem]{ulem}
\useunder{\uline}{\ul}{}

\usepackage{xcolor}
\usepackage{xspace}

\title{EliGen: Entity-Level Controlled Image Generation with Regional Attention}
\author{
Hong Zhang$^{1,2}$\and
Zhongjie Duan$^3$\and
Xingjun Wang$^2$\and
Yingda Chen$^2$\And
Yu Zhang$^{1}$\footnotemark[2] \\
\affiliations
$^1$College of Control Science and Engineering, Zhejiang University\\
$^2$ModelScope Team, Alibaba Group Inc.
$^3$East China Normal University\\
\emails
\{hongzhang99, zhangyu80\}@zju.edu.cn,
zjduan@stu.ecnu.edu.cn, \\
\{xingjun.wxj, yingda.chen\}@alibaba-inc.com
}

\begin{document}
\maketitle

\renewcommand{\thefootnote}{\fnsymbol{footnote}}
\footnotetext[2]{corresponding author}

\begin{abstract}
Recent advancements in diffusion models have significantly advanced text-to-image generation, yet global text prompts alone remain insufficient for achieving fine-grained control over individual entities within an image. To address this limitation, we present EliGen, a novel framework for \textbf{E}ntity-\textbf{l}evel controlled \textbf{i}mage \textbf{Gen}eration. Firstly, we put forward regional attention, a mechanism for diffusion transformers that requires no additional parameters, seamlessly integrating entity prompts and arbitrary-shaped spatial masks. By contributing a high-quality dataset with fine-grained spatial and semantic entity-level annotations, we train EliGen to achieve robust and accurate entity-level manipulation, surpassing existing methods in both spatial precision and image quality. Additionally, we propose an inpainting fusion pipeline, extending EliGen’s capabilities to multi-entity image inpainting tasks. We further demonstrate EliGen’s flexibility by integrating it with other open-source models such as IP-Adapter, In-Context LoRA and MLLM, unlocking new creative possibilities. The source code, model, and dataset are published at the \href{https://github.com/modelscope/DiffSynth-Studio.git}{Project Page}.

\end{abstract}

\section{Introduction}

Recently, diffusion models \cite{rombach2022stablediffusion,esser2024sd3,flux} have significantly advanced text-to-image generation, enabling the creation of high-quality images through textual descriptions. However, a single prompt is insufficient for precise image design, as natural language cannot describe image content with pixel-level precision, and image models often struggle to accurately model spatial relationships and object quantities \cite{podell2023sdxl}. This pitfall significantly limits their broader applicability. Consequently, integrating diverse conditions into text-to-image models has become a central research focus. For example, recent studies like ControlNet \cite{zhang2023controlnet} and IP-Adapter \cite{ye2023ipadapter} have introduced conditions such as depth maps, normal maps and reference images to control the layout and style of generated images. Nevertheless, these conditions operate at a global image level, rather than targeting specific entities within the image. As a result, they lack the capability to provide fine-grained and entity-level control. This limitation impedes the generation of complex, highly customized images, motivating us to investigate precise controlling approaches for entity-level control.


\begin{figure}[tp] 
    \centering
    \includegraphics[width=1\linewidth]{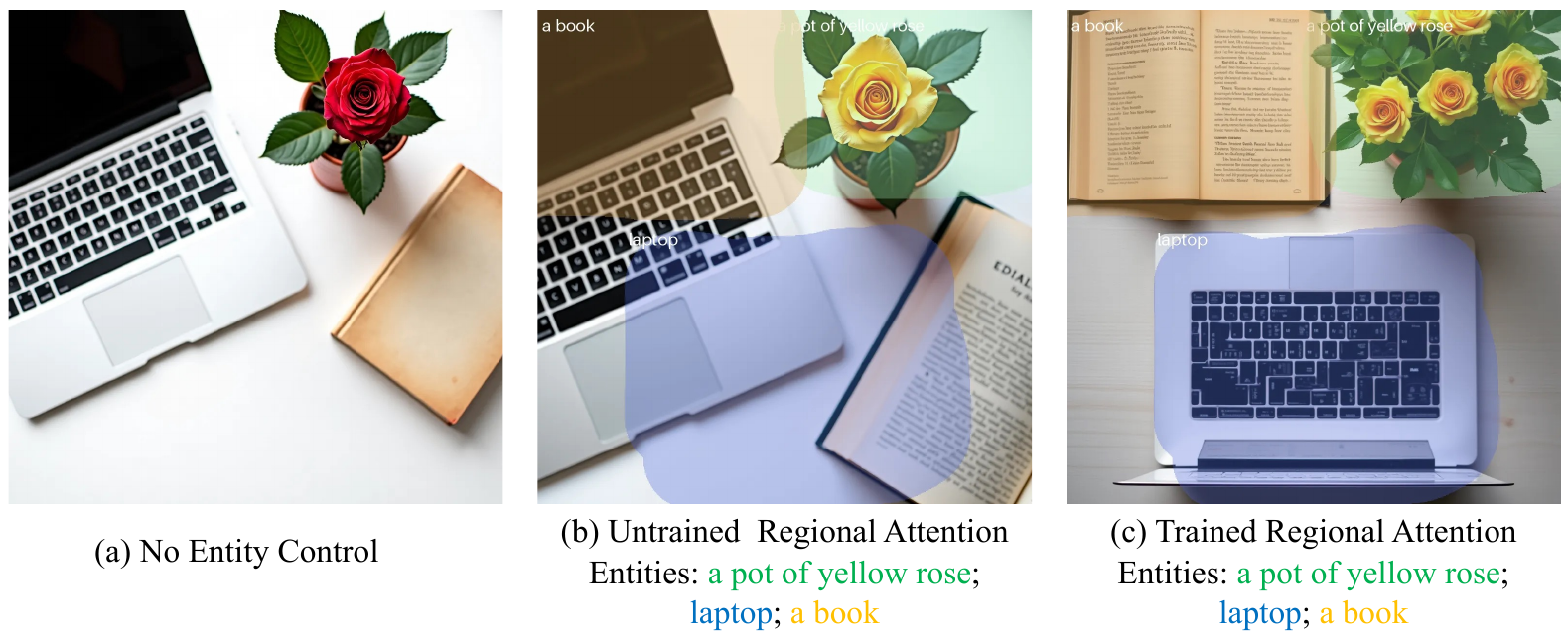}
    \caption{Entity control ability of EliGen. The global prompt is ``top-down view of a desk, laptop, a pot of rose, and a book." (a) No entity control. (b) Untrained regional attention modifies regional details (entity ``rose") but lacks position control ability (entities ``laptop" and ``book"). (c) After training, EliGen successfully achieves control over all entities.}
    \label{fig:train-free}
\end{figure}

\begin{figure*}[t] 
\centering    
\includegraphics[width=1.\linewidth]{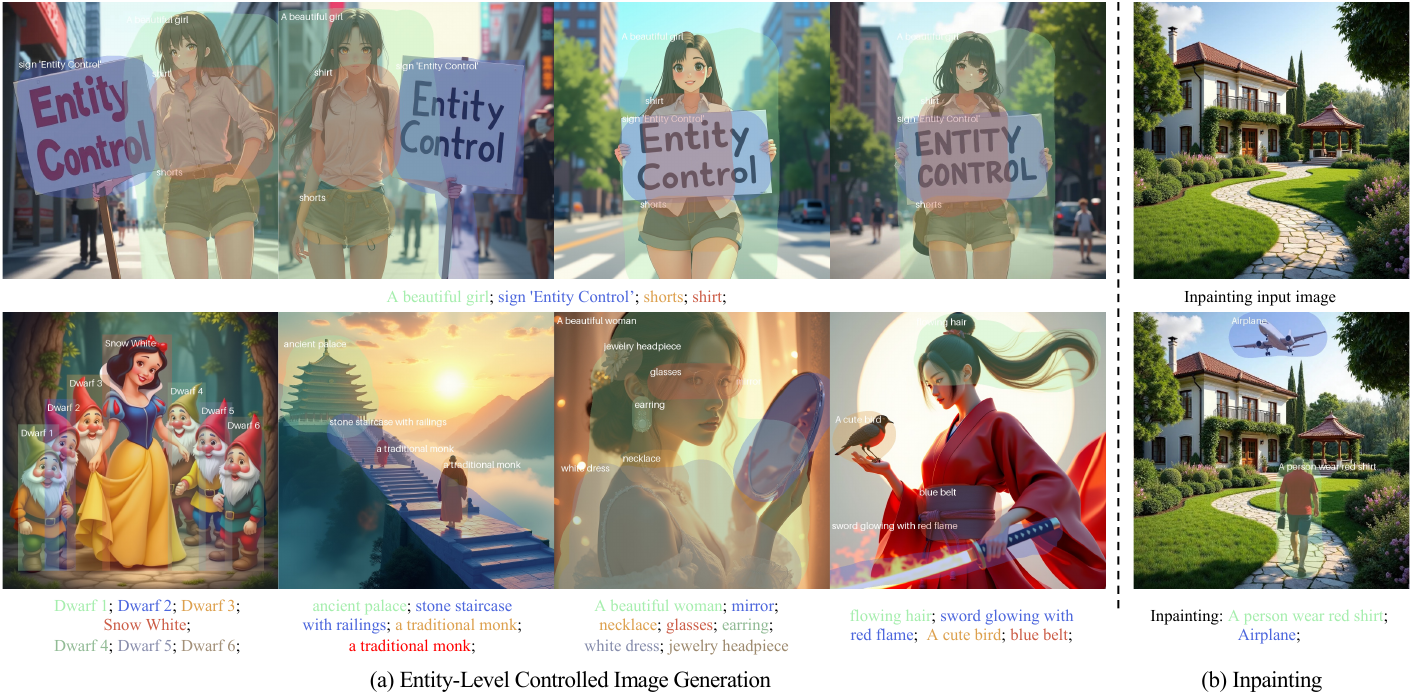}
\caption{EliGen enables spatial and semantic control of each entity. (a) By incorporating local prompts and masks for each entity, it generates images with specific layouts and details. Unlike previous models restricted to rectangular controls, EliGen supports arbitrary-shaped masks, facilitating more creative generation. (b) Additionally, it performs image inpainting with input images. Notably, our model demonstrates robust generalization, consistently producing ideal layouts across different seeds, with detailed experiments in the Supplementary Material.}
\label{fig:cover}
\end{figure*}

In this paper, we focus on adding entity-level control for text-to-image models. Entity-level control is also known as instance-level or open-set layout control. This task allows for the input of multiple entity conditions, each including corresponding spatial and semantic information. Previous studies \cite{li2023gligen,zhou2024migc,wang2024instancediffusion,wu2024ifadapter} have preliminarily demonstrated the feasibility of entity-level control based on the Stable Diffusion series models \cite{rombach2022high,podell2023sdxl}. These methods typically encode spatial conditions in two ways: (1) representing positions via bounding boxes with Fourier encoding \cite{mildenhall2021nerf} and auxiliary modules, or (2) mapping spatial information into attention maps of cross-attention layers. However, these approaches have notable limitations. Firstly, bounding boxes restrict the model's flexibility, as users prefer to input arbitrary-shaped entity masks. Training with precise bounding box coordinates makes the network overly sensitive to numerical inputs, compromising the model's robustness and leading to generated images with subpar aesthetic quality. Secondly, existing methods are closely coupled with the model structure of UNet \cite{ronneberger2015u}, while recent advancements in text-to-image generation have proved that Diffusion Transformers (DiT) \cite{esser2024sd3,flux} are capable of achieving superior performance. Previous methods lack generality and are unable to adapt to the evolving text-to-image models.

Unlike previous approaches that rely on bounding boxes, we aim to generally extend the attention modules of diffusion models to achieve entity-level control with arbitrary-shaped spatial information. To integrate entity conditions into the attention modules, we employ structured text embeddings to represent semantic information, which includes a global prompt describing the overall content of the image and several local prompts describing specific details. Each local prompt is accompanied by a mask representing the corresponding region. Furthermore, based on spatial conditions, we construct joint attention masks (entity-latent, inter-entity, and intra-entity). This enables regional attention for each entity through the extended attention sequences and composed masks. Since local and global prompts share the same embedding space and no additional parameters are introduced, the train-free regional attention mechanism is able to modify entity details. However, it still lacks precise layout control capabilities, as shown in Figure~\ref{fig:train-free}.

Based on the above findings, we further consider training to enhance the model's entity-level control capabilities. For this task, each training sample requires an image accompanied by a global prompt, several local prompts, and corresponding masks. We construct a dataset consisting of 500k high-quality annotated training samples using FLUX.1-dev for image generation and Qwen2-VL \cite{wang2024qwen2vl} for image captioning and entity annotation. Since no additional structures are introduced, we adopt LoRA \cite{hu2021lora} for efficient fine-tuning to ensure faster convergence.

After training, our model, EliGen, acquires precise and robust entity-level control capabilities, seamlessly accepting arbitrary-shaped positional inputs, as shown in Figure~\ref{fig:cover}a and~\ref{fig:train-free}c. Qualitative and quantitative experiments show that EliGen outperforms previous methods in both spatial accuracy and image quality. Furthermore, we introduce a region-based noise fusion operation termed inpainting fusion, enabling EliGen to achieve impressive inpainting effects, as illustrated in Figure~\ref{fig:cover}b. Additionally, EliGen can seamlessly integrate with other open-source models, such as IP-Adapter \cite{flux-ipa}, In-Context LoRA \cite{lhhuang2024iclora}, and MLLM \cite{huang2024context}, to achieve higher levels of creativity, such as styled entity control, entity transfer, and dialogue-based image design and editing.

The contributions of this paper are summarized as follows:
\begin{itemize}
    \item We propose EliGen, a novel approach that leverages fine-grained entity-level information to achieve precise and controllable text-to-image generation.
    \item We present a regional attention mechanism and fine-tune it to focus attention of entites on specific regions, enabling precise entity-level control in EliGen.

    \item Extensive experiments show that EliGen excels in entity-controlled generation. It also performs effectively in image inpainting and seamlessly integrates with community models, proving its strong creative potential.

    \item We release a fine-grained entity-level control dataset to facilitate advanced research in entity-level controlled generation and related tasks.
\end{itemize}

\section{Related Work}

\paragraph{Text-to-Image Diffusion Models.}
Text-to-image diffusion models \cite{DDPM,DDIM,li2024hunyuandit} have catalyzed the evolution of generative AI by generating high-quality images from textual descriptions. Latent Diffusion Models (LDMs) \cite{rombach2022stablediffusion,podell2023sdxl} have emerged as a pivotal innovation, operating in a compressed latent space rather than directly in pixel space, which significantly reduces computational costs. These models primarily incorporate text conditions through the cross attention module within the UNet architecture \cite{ronneberger2015u}. The Diffusion Transformer (DiT) \cite{Peebles2022DiT} represents the first diffusion model to utilize a transformer, integrating conditions into image latents via adaptive LayerNorm. SD3 \cite{esser2024sd3} and FLUX \cite{flux} are the state-of-the-art models that adopt the DiT, enhancing image generation by concatenating text and latent image modalities, thereby fully leveraging the semantic information of text to produce images of higher quality and finer detail. Despite the impressive capabilities of these models, generating images solely from text remains constrained. Recent studies have been exploring methods to introduce additional conditions \cite{zhang2023controlnet,mou2023t2i,ye2023ipadapter}, with entity-level control being one such sophisticated condition.

\paragraph{Entity-Level Controlled Generation.}
To equip foundational models with fine-grained control capabilities, researchers originally explored train-free methods such as multi-diffusion \cite{bar2023multidiffusion} and cross attention guidance \cite{ma2024directed}. However, these approaches often produce unstable results and encounter entity omission issues. To tackle these challenges, researchers introduced additional modules to encode spatial information for each entity using bounding boxes and conducted task-specific training. GLIGEN \cite{li2023gligen} is the first entity control model to undergo training, integrating an additional gated attention module and encoding bounding boxes through Fourier embedding. Subsequently, MIGC \cite{zhou2024migc} introduces the shading aggregation controller and maps spatial information onto attention maps to guide cross attention processes. InstanceDiffusion \cite{wang2024instancediffusion} further categorizes spatial information from coarse to fine and designs unique tokenizers for each category. While these studies focus on designing complex structures to encode spatial information, this paper shifts the focus to designing and training regional attention mechanism without extra parameters and incorporating spatial information by controlling the regions where entity semantics exert influence.

\section{Approach}
\subsection{Preliminaries}
\paragraph{Problem Definition}
Entity-level controlled image generation aims at generating images based on both global and entity-level conditions. The diffusion model should ensure not only the accuracy of attributes and positions of each entity but also the overall harmony and quality of the image. The conditional inputs include a global text prompt $p$ and $L$ entity-level conditions. Each entity $e_i$ is composed of a local text prompt $p_i$ and spatial localization $l_i$, denoted as:
\begin{align}
    \mathrm{Condition}: y &= [p,(e_1,...,e_L)] \\
    \mathrm{Entity}: e_i &= [p_i, l_i]
\end{align}
Here, we adopt the entity mask \( m_i\in \{0,1\}^{h\times w} \) as a general representation for spatial localization, which is a more refined and flexible representation compared to the bounding box \( b_i = \{x_1, y_1, x_2, y_2\} \).

\begin{figure*}[t] 
    \centering
    \includegraphics[width=1\linewidth]{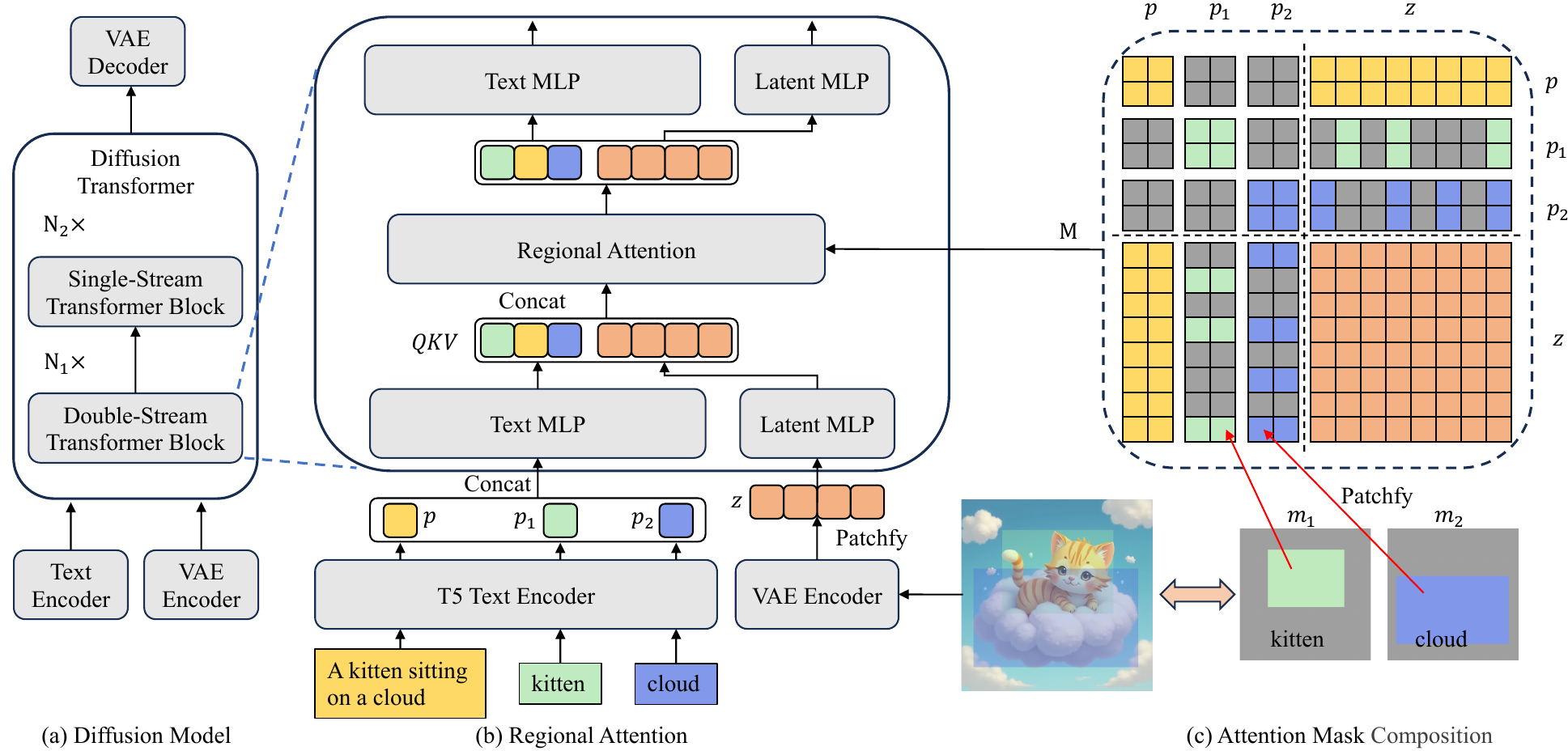}
    \caption{The regional attention mechanism within the double-stream transformer block of DiT. (a) The diffusion model. (b) The global and local prompts are encoded and concatenated with the latent embeddings $z$ to form the attention sequence. (c) The attention mask \( \mathrm{M} \) is constructed from multiple components, each defining the specific region for which each sequence token should perform attention. In the composed mask \( \mathrm{M} \), all colored regions indicate 1, and gray areas indicate 0.}
    \label{fig:regional attention}
\end{figure*}

\paragraph{Motivation}
The core challenge lies in accurately encoding the local prompt \( p_i \) and spatial localization \( l_i \) of each entity and integrating them into the diffusion model. For \( p_i \), integration can be addressed similarly to the global prompt by extracting text embedding with a T5 text encoder \cite{t5}. However, for \( l_i \), prior methods often relied on hard-coding approaches to acquire numerical embedding, such as encoding bounding box coordinates via Fourier embedding or directly using mask images. Such hard-coded methods are suboptimal for the following reasons: \textbf{(1) Rectangular Control Limitation}: Encoding with bounding boxes restricts the model to rectangular control conditions, thereby losing the ability to handle arbitrary shapes, as illustrated in Figure~\ref{fig:cover}a. \textbf{(2) Sensitivity to Spatial Shapes}: Hard-coded numerical values of bounding boxes or masks makes the model sensitive to the coordinates and shapes of entities, which reduces the model's generalizability. \textbf{(3) Additional Structural Requirements}: Hard-coded approaches necessitate the design of extra structures to input localization embeddings.

We utilize a soft-coding scheme to encode \( l_i \). Instead of using numerical spatial inputs, we interpret it as the region where local prompts \( p_i \) exert their influence. Initially, we design a Regional Attention (RA) mechanism to handle each entity condition \( e_i \). While untrained RA can semantically modify regional details, it lacks the ability to precisely control layouts. To address this, we construct a dataset with entity annotations and finetune the model, significantly activating its ability to control layouts effectively.
\subsection{Regional Attention}
This work seeks to extend the attention mechanism to achieve refined control over multiple entities. We adopt the state-of-the-art DiT-based FLUX.1-dev \cite{flux} model as our baseline, while ensuring that the proposed regional attention mechanism can be easily applied to other text-to-image models. The core operation of DiT involves performing joint attention on the text and latent embedding. Given a text embedding $p$ with $n_p$ tokens encoded by the T5 text encoder and a latent embedding $z$ with $n_z$ tokens obtained from a noisy latent image through the patchify operation, DiT concatenates them as a single sequence and do Self-Attention jointly:
\begin{equation}
\label{eq1}
    [z, p] = \mathrm{Attention}([z,p];m)
\end{equation}
In Eq.~\ref{eq1}, $m$ represents the attention mask, which controls the visibility of tokens in the attention operation. In origin FLUX DiT, $m$ is an $N \times N$ dimensional tensor filled with zeros, where $ N = n_p + n_z $, indicating that all tokens are mutually visible to each other.

The core mechanism of regional attention is illustrated in Figure~\ref{fig:regional attention}. The illustration is based on the double-stream transformer block. For the single-stream variant, the overall structure remains the same, except that the Text MLP and Latent MLP share weights. We explain this using an example involving $k=2$ entities. Firstly, as both the global prompt \( p \) and the local prompt \( p_i \) of entity \( e_i \) are text, we use the T5 text encoder to extract their embeddings, with each prompt consisting of $n_p$ tokens. Subsequently, we concatenate them and use a Text MLP to obtain the attention input corresponding to the prompts, which is:
\begin{equation}
\mathrm{H}_p = \mathrm{W_{Text}}([p; p_1; p_2]) + \mathrm{B_{Text}}
\end{equation}
where $\mathrm{H}$ denotes the concatenation of attention inputs: 
\begin{equation}
\mathrm{H}_p = [\mathrm{Q}_p; \mathrm{K}_p; \mathrm{V}_p]
\end{equation}
For the image branch, we use a VAE encoder to extract the latent embeddings, which are then patchified. These embeddings are mapped using a Latent MLP and concatenated with the prompt branch, i.e.,
\begin{align}
& \mathrm{H} = [\mathrm{H}_p; \mathrm{H}_z] \\
& \mathrm{H}_z = \mathrm{W_{Latent}}(z) + \mathrm{B_{Latent}}
\end{align}
Next, the core computation of regional attention is:
\begin{equation}
\mathrm{Attn} = \mathrm{Attention}(\mathrm{H}; \mathrm{M})
\end{equation}
Here, we construct the binary mask $\mathrm{M}$ to specify the regions where each token should attend during attention, as shown in Figure~\ref{fig:regional attention}c, which primarily includes four categories of regions. For consistency, we treat the global prompt as a global ``entity", with its entity mask being 1 for the entire image.

\textbf{Entity-Latent}: The latent embedding \( z \) is derived by patchifying the latent image, where each token corresponds to a specific spatial patch within the image. Leveraging this spatial correspondence, we apply the same patchify operation to transform the entity mask \( m_i \) into a vector that matches the length and positional alignment of \( z \) in Eq.~\ref{eq:patchify}. By applying this vector to each token of the local prompt \( p_i \), we can constrain \( p_i \) to attend exclusively to the designated regions of \( z \).
\begin{equation}
\mathrm{M}(p_i,z) = \mathrm{M}(z,p_i) = \mathrm{patchify}(m_i)
\label{eq:patchify}
\end{equation}

\textbf{Inter-Entity}: In baseline DiT, there was only the one entity with global prompt \( p \). After introducing local prompts, each entity should not be visible to others to avoid semantic information leakage between entities, i.e.,
\begin{equation}
\mathrm{M}(p_i,p_j) = \mathrm{M}(p_j,p_i) = 0 \quad \mathrm{if} \ i \neq j
\label{eq:regional_attn_0}
\end{equation}

\textbf{Intra-Entity}: Since we have adopted joint attention, the prompt embedding $p_i$ should also incorporate its own information during attentional updates, i.e.,
\begin{equation}
\mathrm{M}(p_i,p_i) = 1
\end{equation}

\textbf{Latent-Latent}: Similar to prompt embedding, the latent embedding \( z \) should also be visible to itself, i.e.,
\begin{equation}
\mathrm{M}(z,z) = 1
\end{equation}

After constructing the above regions, we need to concatenate them together to form a $N_r \times N_r$ attention map $\mathrm{M}$ as Figure~\ref{fig:regional attention}c, where $N_r = (k+1) * n_p + n_z$. With \( \mathrm{M} \), the regional attention mechanism defined in Eq.~\ref{eq:regional_attn_0} can be expressed in detail as:



\begin{equation}
\mathrm{Attn}=\mathrm{Softmax} \left (\frac{\mathrm{Q}\mathrm{K^{T}}}{\sqrt{d}}+ \mathrm{log(M)}\right ) V
\end{equation}

Compared to the baseline, the \( p_i \) we introduce shares the same data distribution as \( p \). Consequently, the model with regional attention can generate harmonious images without training. However, training is still essential to acquire the ability to control layout. To this end, we contribute a dedicated entity control dataset.
\subsection{Dataset with Entity Annotation}

For the entity control task, each training sample should consist of an image, a global prompt, and multiple entity annotations. Regarding the image source, prior works typically use open-source datasets like Laion \cite{LAION-Aesthetics} for images, but we observe that Laion's image distribution differs significantly from images generated by base model, leading to stylistic inconsistencies. Using Laion for training would distort the model distribution and degrade image quality (see the Supplementary Material for ablations), as the model would need to fit this mismatch. Therefore, we choose to directly use images generated by our baseline model, FLUX.1-dev, to avoid such distribution discrepancies.

The dataset generation pipeline and statistics are illustrated in the Supplementary Material. Starting with DiffusionDB \cite{wangDiffusionDBLargescalePrompt2022}, we randomly select a caption and use FLUX.1-dev to generate the source image. Then, we employ Multimodal Large Language Model (MLLM) \cite{wang2024qwen2vl} to re-caption the image, forming the global prompt. Unlike previous methods relying on Grounding DINO \cite{liu2023grounding}, we directly use Qwen2-VL 72B, which possesses the most powerful grounding capabilities across all MLLMs, to randomly select entities in the image and annotate their local prompts and localization (via rectangular masks). This ensures consistency between local and global prompts. Since our regional attention employs soft-coded spatial localization, training with rectangular masks is sufficient to enable the model to generalize to arbitrary-shaped masks.

\subsection{Implementation Details}
Our model does not introduce any additional parameters. To achieve faster convergence and better alignment with the community, we utilize the LoRA (Low-Rank Adaptation) \cite{hu2021lora} method for fine-tuning. The LoRA weights are applied to the linear layers of each block in DiT, including the projection layers before and after attention, as well as the linear layers within adaptive LayerNorm. The LoRA Rank is set to 64. The model is trained with a batch size of 64 for 20K steps using the AdamW optimizer and a learning rate of 0.0001. To minimize memory usage, we employ the DeepSpeed \cite{rasley2020deepspeed} training strategy with a bfloat16 precision. The model is trained with the standard generation loss of rectified flow \cite{esser2024sd3}. During inference, we utilize the Euler scheduler \cite{karras2022elucidating} with 50 sampling steps and set the classifier-free guidance \cite{ho2022classifier} scale to 3. Additionally, the embedded guidance within FLUX is configured to 3.5.

\section{Qualitative Experiments}
\begin{figure*}[t] 
    \centering
    \includegraphics[width=1.\linewidth]{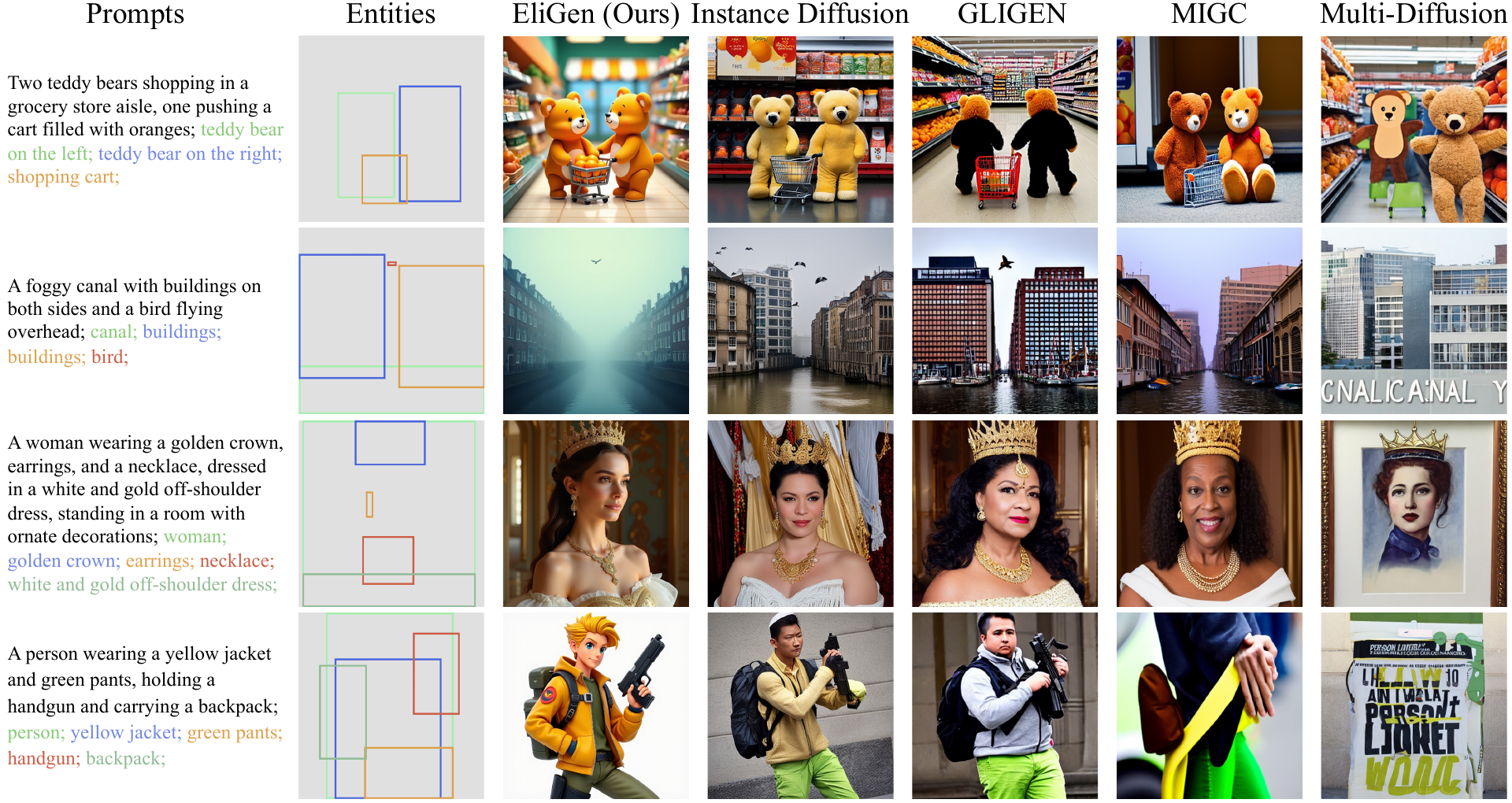}
    \caption{Qualitative results conditioned on multiple rectangular-shaped entities. Test case combinations evolve from simple to complex, with the final two rows illustrating enhanced graph quality and coherence of our model in the presence of entity coupling.}
    \label{fig:bbox_result}
\end{figure*}

\begin{figure}[t] 
    \centering
    \includegraphics[width=0.95\linewidth]{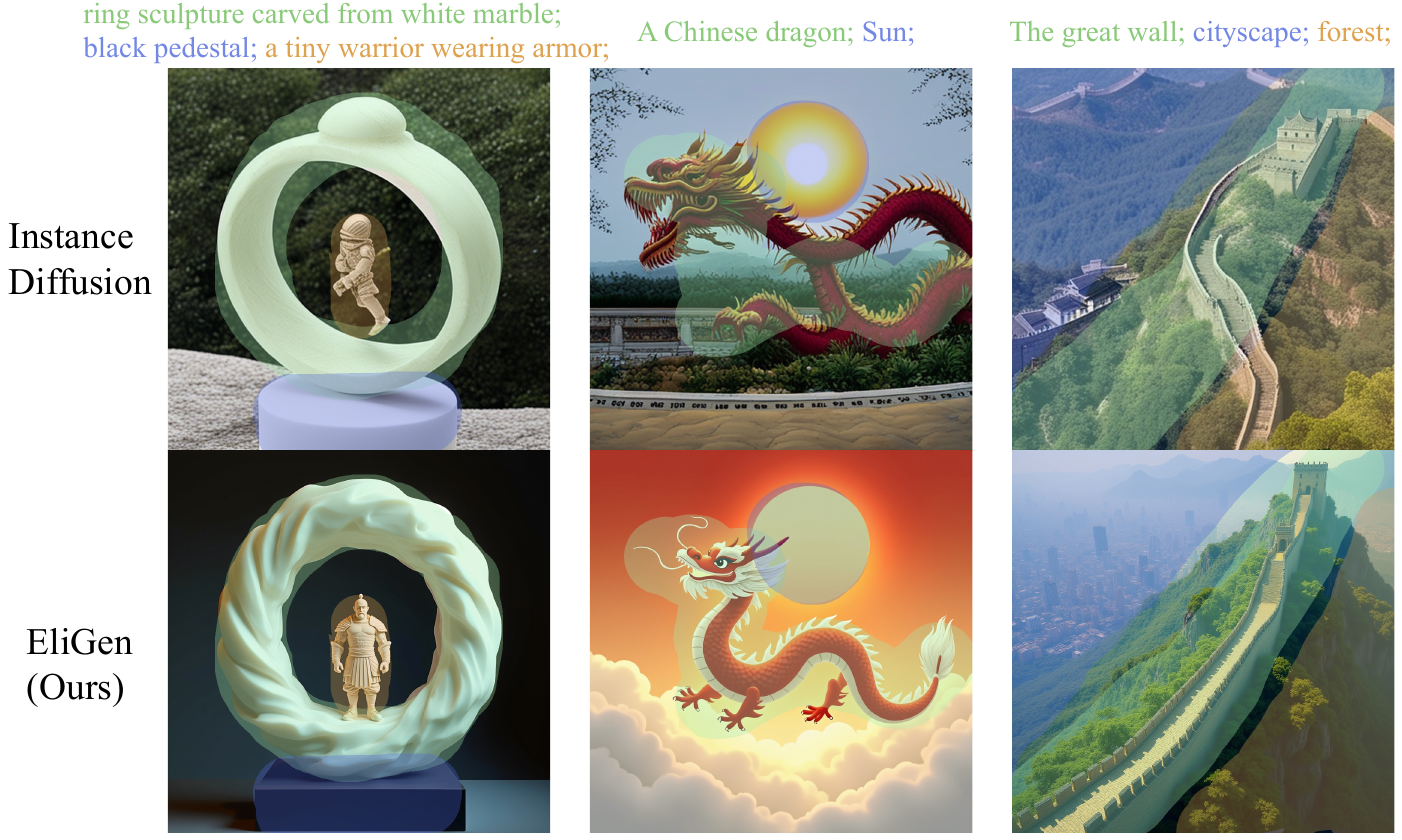}
    \caption{Qualitative results with arbitrary-shaped entities.}
    \label{fig:mask_result}
\end{figure}

\subsection{Entity-Level Controlled Generation}
We assess our model's generalization in the entity control task by evaluating it under two masked input scenarios and comparing it with prior works including Instance Diffusion \cite{wang2024instancediffusion}, GLIGEN \cite{li2023gligen}, MIGC \cite{zhou2024migc}, and Multi-Diffusion \cite{bar2023multidiffusion}.
\paragraph{Rectangular Masks as Input}
In this scenario, entities are localized using bounding boxes. Our model converts bounding boxes to masks for input, while others utilize bounding box coordinates. The results are shown in Figure~\ref{fig:bbox_result}, where our model demonstrates superior performance in image quality, and entity accuracy. In the first two rows, simple tasks with distinct entities are handled by most methods except Multi-Diffusion. The last two rows highlight complex layouts where entities are intricately coupled. While other models struggle to maintain accuracy and coherence, our model consistently delivers high-quality and coherent images.
\paragraph{Arbitrary Masks as Input}
We further conduct experiments to verify the generalization ability of our method for arbitrary-shaped masks. Prior to this work, Instance Diffusion was the sole approach capable of handling arbitrary-shaped masks as input. The results, illustrated in Figure~\ref{fig:mask_result}, clearly show that EliGen outperforms Instance Diffusion in image quality, positional accuracy, and attribute fidelity. Instance Diffusion struggles with position control for masks that significantly deviate from rectangular shapes, such as ``Chinese dragon" and ``Great Wall". In contrast, EliGen consistently delivers excellent results.


\subsection{Image Inpainting}
Image inpainting targets rendering desired content at specified locations. While encoding the input image as the initial latent for the EliGen model is intuitive, it often disrupts non-inpainting regions. Thus, we propose the inpainting fusion pipeline to preserve these areas while enabling precise entity-level modifications over inpaining regions. The core mechanism involves integrating entity latents from inpainting regions with background latents from non-inpainting areas. Due to space constraints, the algorithmic details are provided in the Supplementary Material.

\begin{figure}[t] 
    \centering
    \includegraphics[width=0.95\linewidth]{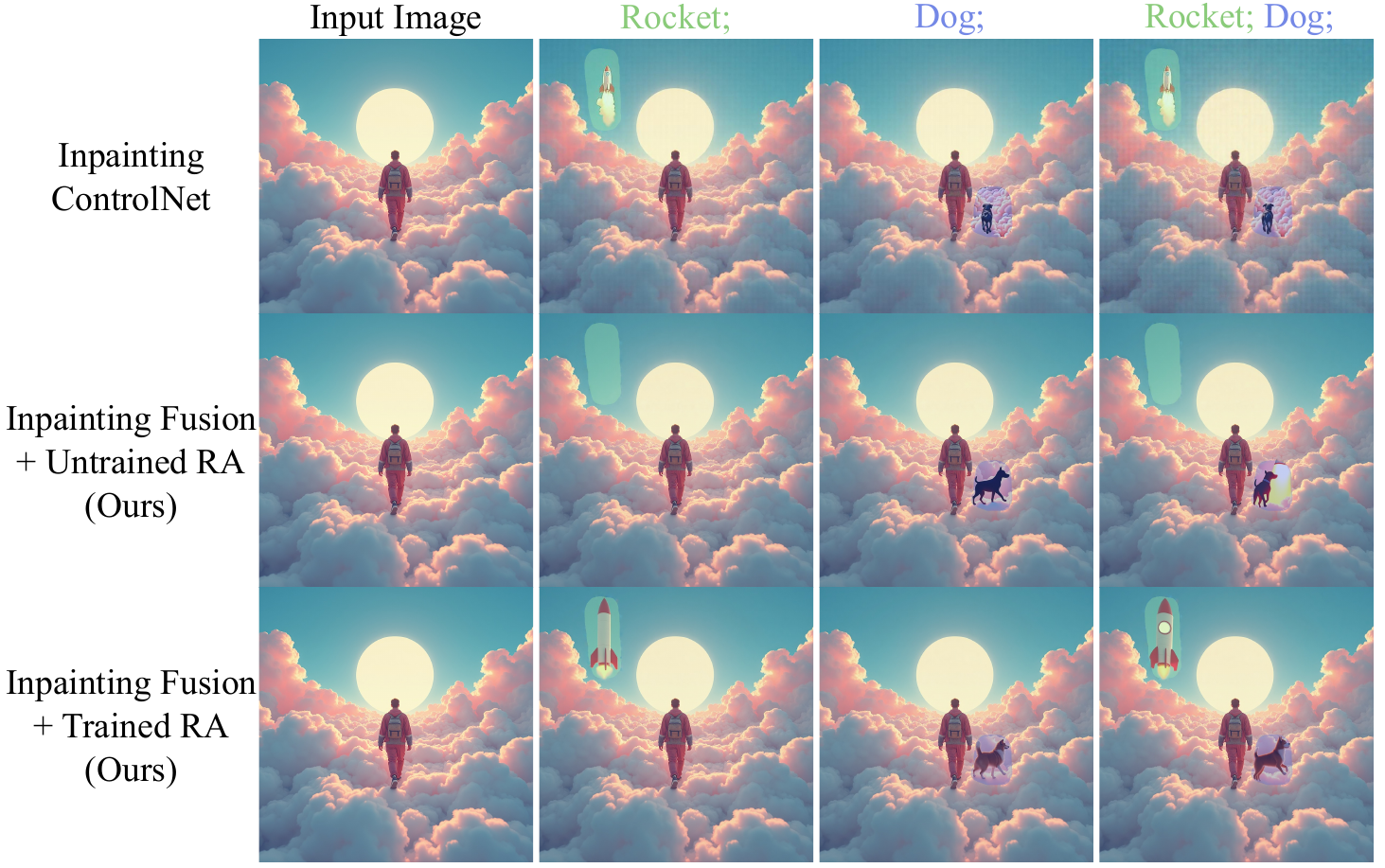}
    \caption{Qualitative comparison on image inpainting. Global prompt for all methods are generated by MLLM.}
    \label{fig:inpainting}
\end{figure}

We compare our method with FLUX-ControlNet \cite{FLUX-Controlnet} in Figure~\ref{fig:inpainting}. As is shown in the result, inpainting fusion ensures background invariance, and training regional attention further enables EliGen to render desired objects in the inpainting region. Compared to FLUX-ControlNet, our model excels at: (1) preserving non-inpainting areas without quality loss, (2) enabling precise control over entity position and shape for harmonious blending, and (3) supporting multi-entity inpainting in a single forward pass.

\begin{table*}[]
\centering
\caption{Evaluation on COCO benchmark from multiple dimensions: Entity Success Rate, Spatial Accuracy and Image Quality. $\uparrow$ indicates that a higher value is better. \textbf{Bold} and \uline{underline} denote the best and second best methods, respectively.}
\label{tab:tb1}
\begin{tabular}{cccccccc}
\toprule
\multirow{2}{*}{{Model}} &
  \multicolumn{3}{c}{{Entity Success Rate (\%)}} &
  {\begin{tabular}[c]{@{}c@{}}Spatial\\ Accuracy (\%)\end{tabular}} &
  \multicolumn{3}{c}{{Image Quality}} \\ \cmidrule(lr){2-4} \cmidrule(lr){5-5} \cmidrule(lr){6-8}  
 &
  {Qwen2VL $\uparrow$} &
  {InternVL2 $\uparrow$} &
  {CogVLM2 $\uparrow$} &
  {mIoU $\uparrow$} &
  {\begin{tabular}[c]{@{}c@{}}Entity\\ CLIP $\uparrow$\end{tabular}} &
  {\begin{tabular}[c]{@{}c@{}}Global\\ CLIP $\uparrow$\end{tabular}} &
  {\begin{tabular}[c]{@{}c@{}}Aesthetic\\ Score $\uparrow$\end{tabular}} \\ \midrule
Multi-Diffusion    & 51.17          & 50.38          & 83.15          & 31.74          & 20.07          & 22.58          & 5.04          \\
GLIGEN             & 71.83          & 68.47          & 88.51          & 59.48          & 20.15          & {\ul 24.94}    & {\ul 5.07}    \\
MIGC               & 77.01          & 74.48          & 89.76          & 64.75          & 20.48          & 23.57          & 5.06          \\
Instance Diffusion & {\ul 82.43}    & {\ul 79.49}    & {\ul 91.33}    & \textbf{76.17} & {\ul 21.06}    & 24.17          & 4.89          \\ \midrule
\textbf{EliGen (Ours)}      & \textbf{88.41} & \textbf{83.79} & \textbf{96.25} & {\ul 73.93}    & \textbf{21.47} & \textbf{27.39} & \textbf{5.60} \\ \bottomrule
\end{tabular}
\end{table*}

\begin{table}[]
\centering
\caption{Human preference evaluation on COCO benchmark. The numerical values represent the win rate in pairwise comparisons. All scores significantly exceeding 50\% demonstrate that EliGen aligns more closely with human preferences compared to other methods.}
\label{tab:tb2}
\begin{tabular}{@{}ccc@{}}
\toprule
Comparison                    & MPS (\%) & PickScore (\%) \\ \midrule
EliGen vs. Instance Diffusion & 71.19    & 65.70          \\
EliGen vs. Gligen             & 60.71    & 60.55          \\
EliGen vs. MIGC               & 74.44    & 69.48          \\
EliGen vs. Multi-Diffusion    & 69.60    & 72.32          \\ \bottomrule
\end{tabular}
\end{table}

\subsection{Potential Creative Extension}
EliGen holds boundless creative potential, and we demonstrate its capabilities when integrated with IP-Adapter \cite{flux-ipa}, In-Context LoRA \cite{lhhuang2024iclora} and MLLM \cite{wang2024qwen2vl}. Due to space constraints, we provide detailed methods and results in the Supplementary Material, and we highly recommend reviewing them.

\section{Quantitative Experiments}

\subsection{Experimental Setup}
Following previous methods, we perform quantitative evaluations on the filtered COCO \cite{lin2014coco,zheng2023layoutdiffusion} 2017 validation set, comprising 2830 prompt-image pairs. Previous GLIGEN \cite{li2023gligen} utilized the mean average precision for evaluation, a metric that computes the area under the precision-recall curve. However, we observe its critical limitation: it treats background entities in generated images as negative samples, resulting in substantial inaccuracies. As a result, we decide not to use this metric. Instead, to thoroughly evaluate the models' capabilities on the COCO benchmark, we assess their performance across the following dimensions.
\paragraph{Entity Success Rate}
To evaluate the model's ability to generate entities accurately, we assess the quality of the target regions, entity details, and image-text alignment with MLLMs. The success rate is then calculated by aggregating results across all entities in the benchmark.
\paragraph{Spatial Accuracy}
We evaluate the spatial accuracy of generated entities by calculating the mean Intersection-over-Union (mIoU) between the ground truth bounding boxes and the predicted bounding boxes. The predicted bounding boxes are obtained using YOLOv11x \cite{khanam2024yolov11}.
\paragraph{Image Quality}We evaluate semantic accuracy using the CLIP score for both entities and the global image. Additionally, Aesthetic Score is adopted to measure the visual quality.
\paragraph{Human Preference}To assess the alignment with human preferences, we compare EliGen against other models using two metrics: the Multi-dimensional Preference Score (MPS) \cite{MPS} and PickScore \cite{Kirstain2023PickaPicAO}.

\subsection{Comparison on COCO Benchmark}
The evaluation results of the models on the COCO benchmark are presented in Table~\ref{tab:tb1}. For Entity Success Rate, we assessed all methods using Qwen2-VL \cite{wang2024qwen2vl}, InternVL2 \cite{chen2024internvl}, and CogVLM2 \cite{hong2024cogvlm2}. The results demonstrate that EliGen consistently outperforms other models, highlighting its robustness and high quality. In terms of mIoU, our model performs comparably to Instance-Diffusion and substantially better than other methods. The slightly lower mIoU of EliGen compared to Instance-Diffusion is primarily due to its soft-coding scheme for positional information. This scheme allows EliGen to automatically expand certain regions to ensure image coherence, as further illustrated in the Supplementary Material. In terms of both CLIP Score and Aesthetic Score, EliGen significantly outperforms other models, demonstrating substantial superiority in image quality.

The human preference results in Table~\ref{tab:tb2} measure the probability of humans favoring one image over another in pairwise comparisons. EliGen consistently achieves a preference probability well above 50\% against all other models, demonstrating its stronger alignment with human preferences and higher image quality of generated images.

\begin{table}[t]
\caption{Results of user study. Each user was asked to rate an image across the four dimensions on a scale from 1 to 5.}
\label{tab:tb3}
\scalebox{0.9}{
\begin{tabular}{@{}ccccc@{}}
\toprule
Model &
  \begin{tabular}[c]{@{}c@{}}Spatial\\ Accuracy\end{tabular} &
  \begin{tabular}[c]{@{}c@{}}Entity\\ Details\end{tabular} &
  \begin{tabular}[c]{@{}c@{}}Entity\\ Coherence\end{tabular} &
  \begin{tabular}[c]{@{}c@{}}Image\\ Aesthetic\end{tabular} \\ \midrule
Multi-Diffusion    & 2.90          & 2.49          & 2.53          & 2.38          \\
GLIGEN             & 4.23          & 3.24          & {\ul 3.53}    & 3.28          \\
MIGC               & 4.15          & 2.82          & 3.09          & 2.79          \\
Instance Diffusion & {\ul 4.45}    & {\ul 3.37}    & 3.51          & {\ul 3.31}    \\
\textbf{EliGen (Ours)}    & \textbf{4.62} & \textbf{4.44} & \textbf{4.31} & \textbf{4.30} \\ \bottomrule
\end{tabular}
}
\end{table}
\subsection{User Study}
To complement the automated metrics, we conducted a user study for all models. We randomly select 250 samples with $1\sim3$ entity conditions from the COCO benchmark and invite users to evaluate these examples across multiple dimensions, including Spatial Accuracy, Entity Details, Entity Coherence, and Image Aesthetic. The Entity Coherence metric assesses how well the entities integrate with other entities or the environment, ensuring that no isolated entities appear in the generated images.

We collected a total of 5,500 ratings, with the evaluation results presented in Table~\ref{tab:tb3}. EliGen achieves the highest scores across all four dimensions, particularly excelling in Entity Details, Entity Coherence, and Image Aesthetic. These results are consistent with those in Table~\ref{tab:tb1}, demonstrating EliGen's robustness in controlling entities with higher spatial precision and superior image quality.

\section{Conclusion}
To achieve image generation with entity-level controls, we introduce regional attention into the DiT architecture to effectively integrate entity prompts and their locations. Furthermore, we construct a dataset with fine-grained entity information and finetune the regional attention mechanism, developing the EliGen model. EliGen demonstrates high generalization in entity-level controlled image generation, delivering superior image quality and entity accuracy compared to existing models. We further propose an inpainting fusion pipeline, extending EliGen’s functionality to image inpainting tasks. Moreover, by integrating EliGen with open-source models, we demonstrate its potential for creative applications, broadening its scope and versatility.

\bibliographystyle{named}
\bibliography{ijcai25}

\appendix
\clearpage
\setcounter{page}{1}
\renewcommand{\title}[1]{\titleold{#1}\newcommand{\thetitle}{#1}}

\newpage
       \twocolumn[
        \centering
        \Large
        \textbf{EliGen: Entity-Level Controlled Image Generation with Regional Attention}\\
        \vspace{0.5em}Supplementary Material \\
        \vspace{1.0em}
       ]

\section{Dataset Details}

\subsection{Dataset Construction Pipeline}
\label{sec:pipeline}
\begin{figure}[h] 
    \centering
    \includegraphics[width=1\linewidth]{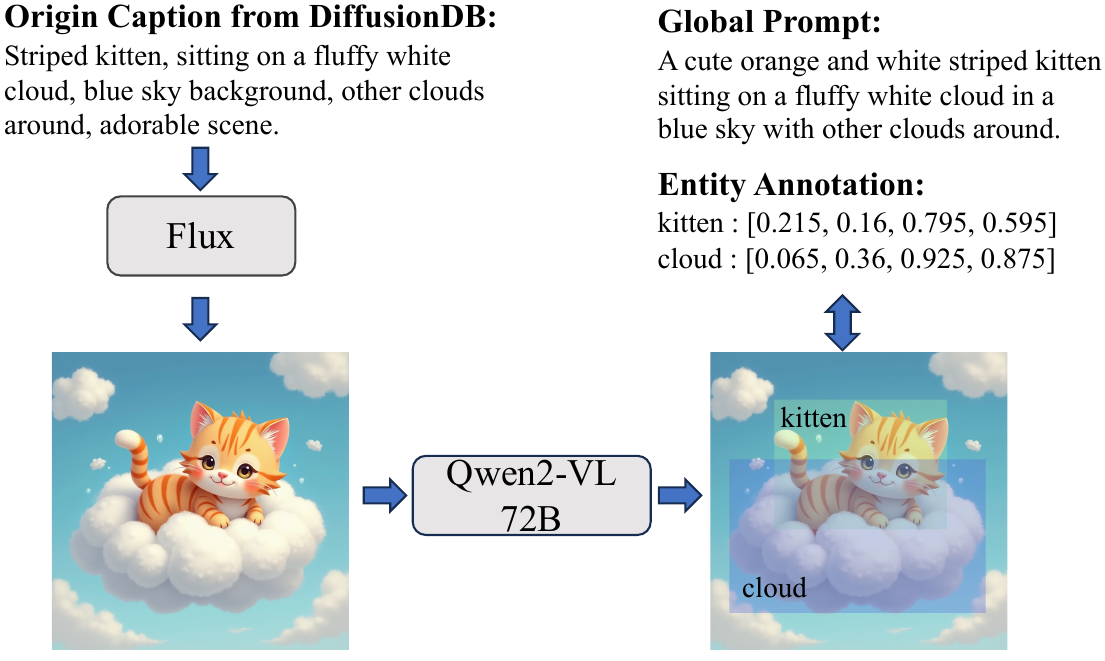}
    \caption{The dataset generation and annotation pipeline. We utilize Flux to generate images and annotate them with Qwen2-VL.}
    \label{fig:dataset pipelie}
\end{figure}
As illustrated in Figure~\ref{fig:dataset pipelie}, to circumvent the additional training overheads that may arise from data distribution differences, we leverage FLUX.1-dev \cite{flux} for the direct generation of training images. Following this, we deploy Qwen2-VL 72B \cite{wang2024qwen2vl} to annotate the images. This process involves crafting a comprehensive global description of the image, identifying entities present within the image, and meticulously annotating each entity with textual descriptions and their respective spatial locations. Ultimately, we create approximately 500K training samples with 1.27M entity annotations.

\begin{figure}[t] 
    \centering
    \includegraphics[width=1.\linewidth]{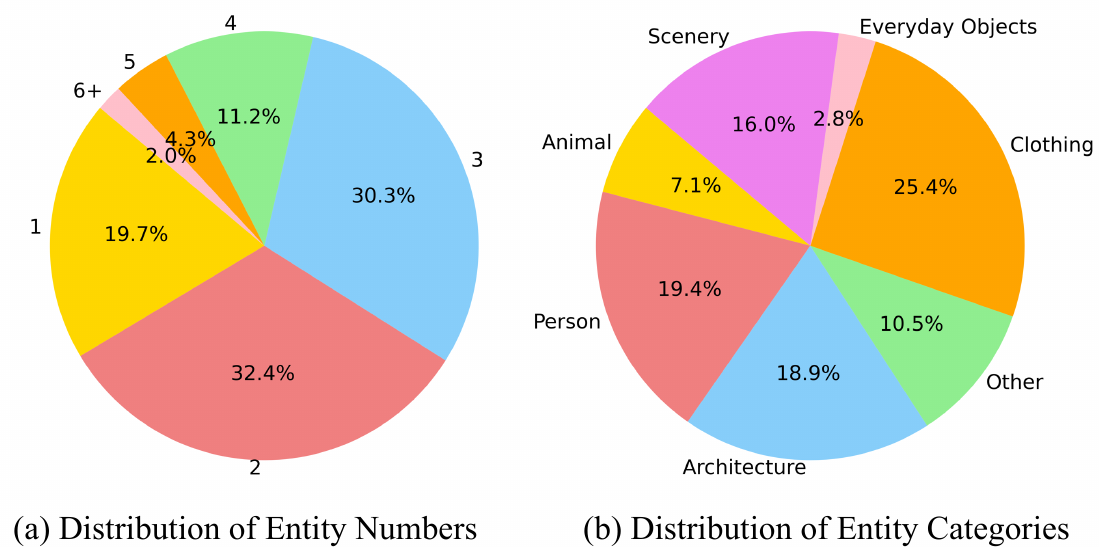}
    \caption{The dataset distributions of entity numbers and categories.}
    \label{fig:dataset distribution}
\end{figure}

\subsection{Dataset Statistics}
We conduct a statistical analysis of the annotated dataset in Figure~\ref{fig:dataset distribution}, primarily focusing on the distribution of the number of entities per image and the semantic categories of the entities.

In terms of the number of entities, the vast majority of samples contain 1 to 5 entities, while approximately 2\% of the samples have more than 6 entities. This distribution of entities is training-friendly because a stable number of entities ensures that the memory usage and computational resource consumption per training step remain relatively consistent. Additionally, since our designed regional attention mechanism supports a variable number of entities, the trained EliGen model demonstrates strong generalization capabilities even for cases with more than 6 entities, effectively handling multi-entity scenarios.

Regarding entity categories, our dataset demonstrates substantial diversity, with entities broadly classified into 7 categories. Since annotations are open-set (not limited to predefined categories), we use Qwen2.5 3B \cite{qwen2.5} to categorize them into the following groups for statistical analysis:

\begin{itemize}
    \item Person: human beings, including men, women, children, and fictional characters.
    \item Scenery: natural landscapes, such as mountains, forests, rivers, and beaches.
    \item Architecture: buildings, structures, and man-made environments.
    \item Clothing: apparel and accessories worn by people.
    \item Animal: living creatures, including pets and wild animals.
    \item Everyday Objects: common items used in daily life, such as furniture, tools, and gadgets.
    \item Other: all entities that do not fit into the above categories, including fictional elements.
\end{itemize}

\begin{algorithm*}[htp]
    \caption{Image Inpainting Fusion Algorithm}
    \label{alg:inpainting}
\textbf{Input:} Input image $I$, inpainting masks $\{m_1, \dots, m_n\}$, global prompt $p$, local prompts $\{p_1, \dots, p_n\}$, negative prompt $p_{neg}$, and hyperparameters $\{\sigma_t\}_{t=1}^T$, $cfg$; VAE model \{$\mathrm{VAE}=(E(I), D(z)$\} and Diffusion Transformer \{$\mathrm{DiT}(z, p, m, t)$\}\\
\textbf{Output:} Inpainted image $P$ with entities $\{p_1, \dots, p_n\}$ inpainted on regions $\{m_1, \dots, m_n\}$ \\
\begin{algorithmic}
\STATE Encode the input image: $z_{init} \gets E(I)$
\STATE Initialize $z_T$ with random noise: $z_T \sim \text{noise}(z_{init}, T)$
\FOR{each denoising step $t = T, \dots, 0$}
    \STATE Compute foreground noise: $Nf_t \gets \mathrm{DiT}(z_t; p; \{p_1, \dots, p_n\}; \{m_1, \dots, m_n\}; t)$
    \STATE Compute background noise: $Nb_t \gets (z_t - z_{init}) / \sigma_t$
    \STATE Fuse foreground and background noise: $Npos_t \gets Nf_t \cdot \cup m_i + Nb_t \cdot (1 - \cup m_i)$
    \STATE Compute negative noise: $Nneg_t \gets \mathrm{DiT}(z_t; p_{neg}; t)$
    \STATE Apply classifier-free guidance: $N_t \gets Nneg_t + cfg \cdot (Npos_t - Nneg_t)$
    \STATE Update latent embedding with flow matching pipeline: $z_{t-1} \gets z_t + N_t \cdot (\sigma_{t-1} - \sigma_t)$
\ENDFOR
\STATE Decode the denoised latents: $P=D(z_0)$
\STATE \textbf{Return:} Inpainted image $P$
\end{algorithmic}
\end{algorithm*}

\begin{figure*}[h]
    \centering
    \includegraphics[width=0.95\linewidth]{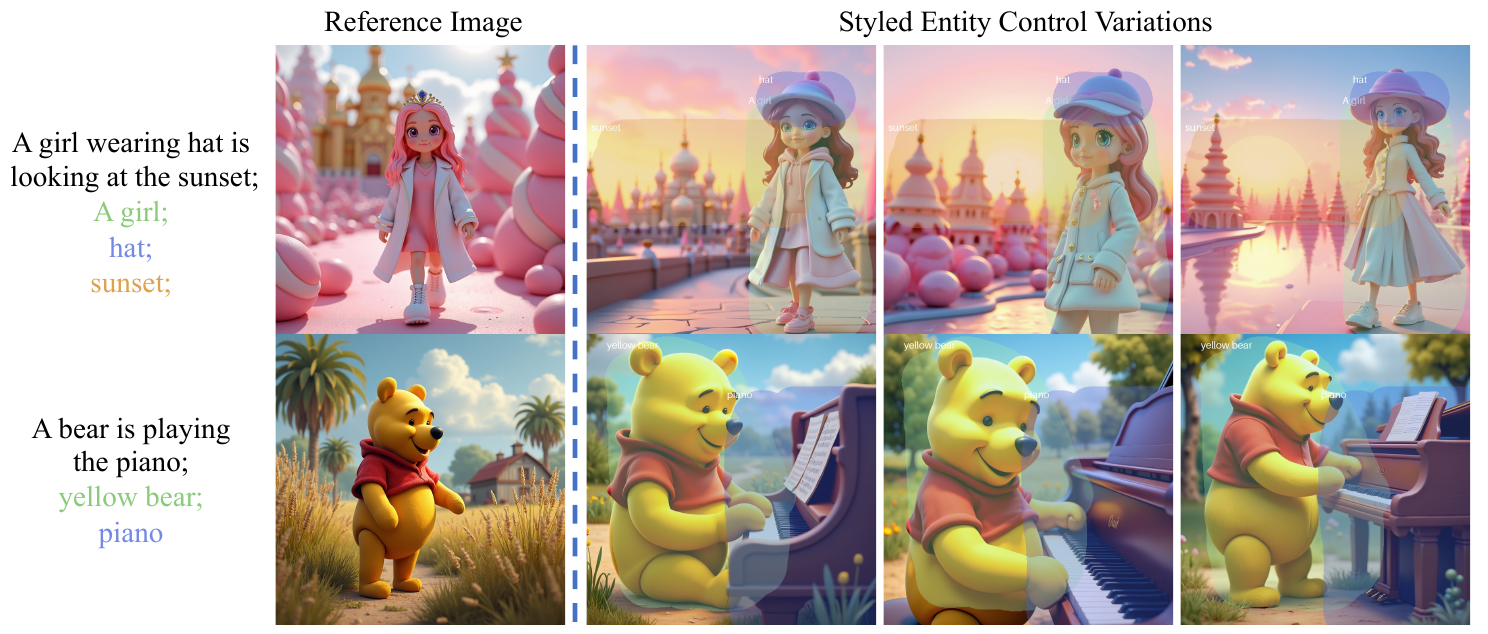}
    \caption{Styled entity control with IP-Adapter and our EliGen. Upon receiving the desired entity information, we can generate a target image that mirrors the style and characters of the reference image, while ensuring the position and attributes of the entities are precisely controlled.}
    \label{fig:IP-Adapter}
\end{figure*}

\section{Inpainting Fusion Algorithm}
Image inpainting aims to reconstruct or modify specific regions of an input image, such as adding or replacing objects. To leverage the entity-level generation capabilities of EliGen for image inpainting while ensuring that non-inpainting regions remain unchanged and the inpainting regions align with the overall image style, we propose the Inpainting Fusion Algorithm, as detailed in Algorithm~\ref{alg:inpainting}. The input image $I$ is first encoded into an initial latent embedding $z_{init}$. At each denoising step, the predicted noise from the DiT is fused with the noise derived from the original image. This fusion ensures that the foreground regions are inpainted with the generated entities, while the background regions remain consistent with the original image. Furthermore, we incorporate classifier-free guidance to refine the overall image quality during the fusion process. After \( T \) denoising steps, the denoised latent embedding $z_0$ is decoded to produce the final inpainted image $P$. This approach achieves precise entity-level modifications while maintaining the coherence and stylistic consistency of the entire image.

\begin{figure*}[thp] 
    \centering
    \includegraphics[width=1.\linewidth]{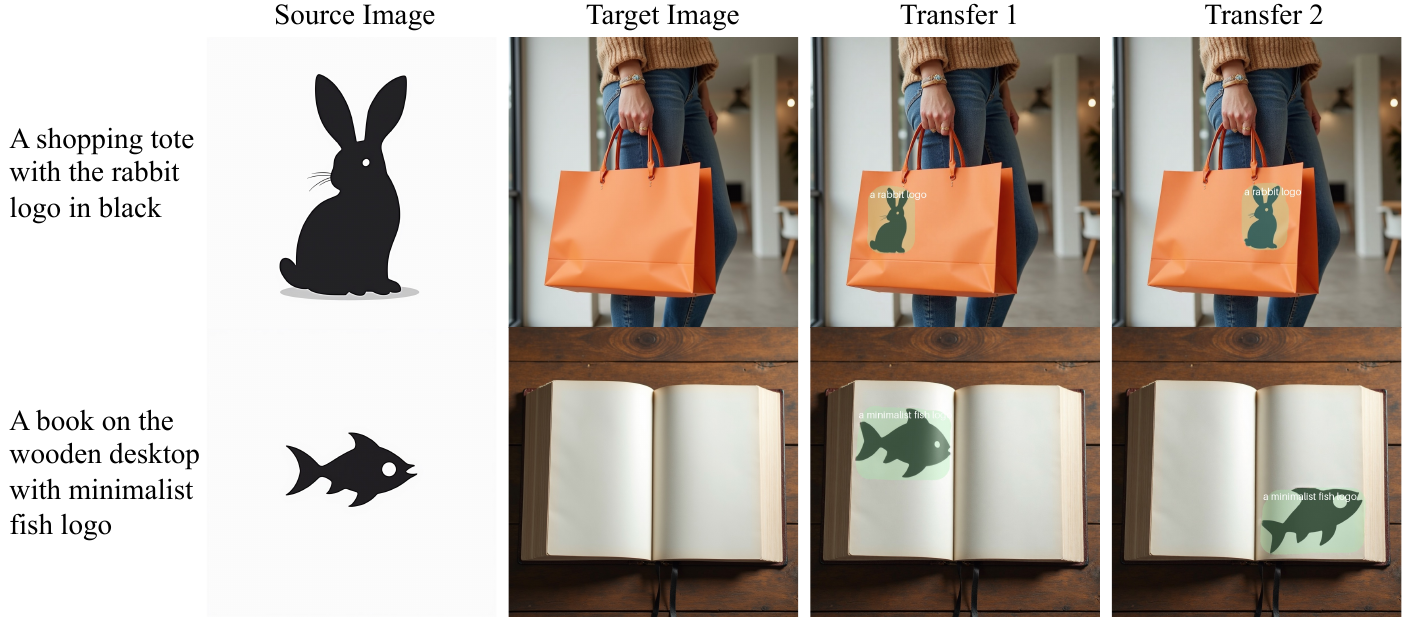}
    \caption{Entity transfer with In-Context LoRA and our EliGen. Given the identity defined by the source image, EliGen can seamlessly transfer it to any desired location and scale within the target image, thereby achieving the ideal visual design.}
    \label{fig:iclora}
\end{figure*}

\begin{figure*}[thp] 
    \centering
    \includegraphics[width=1.\linewidth]{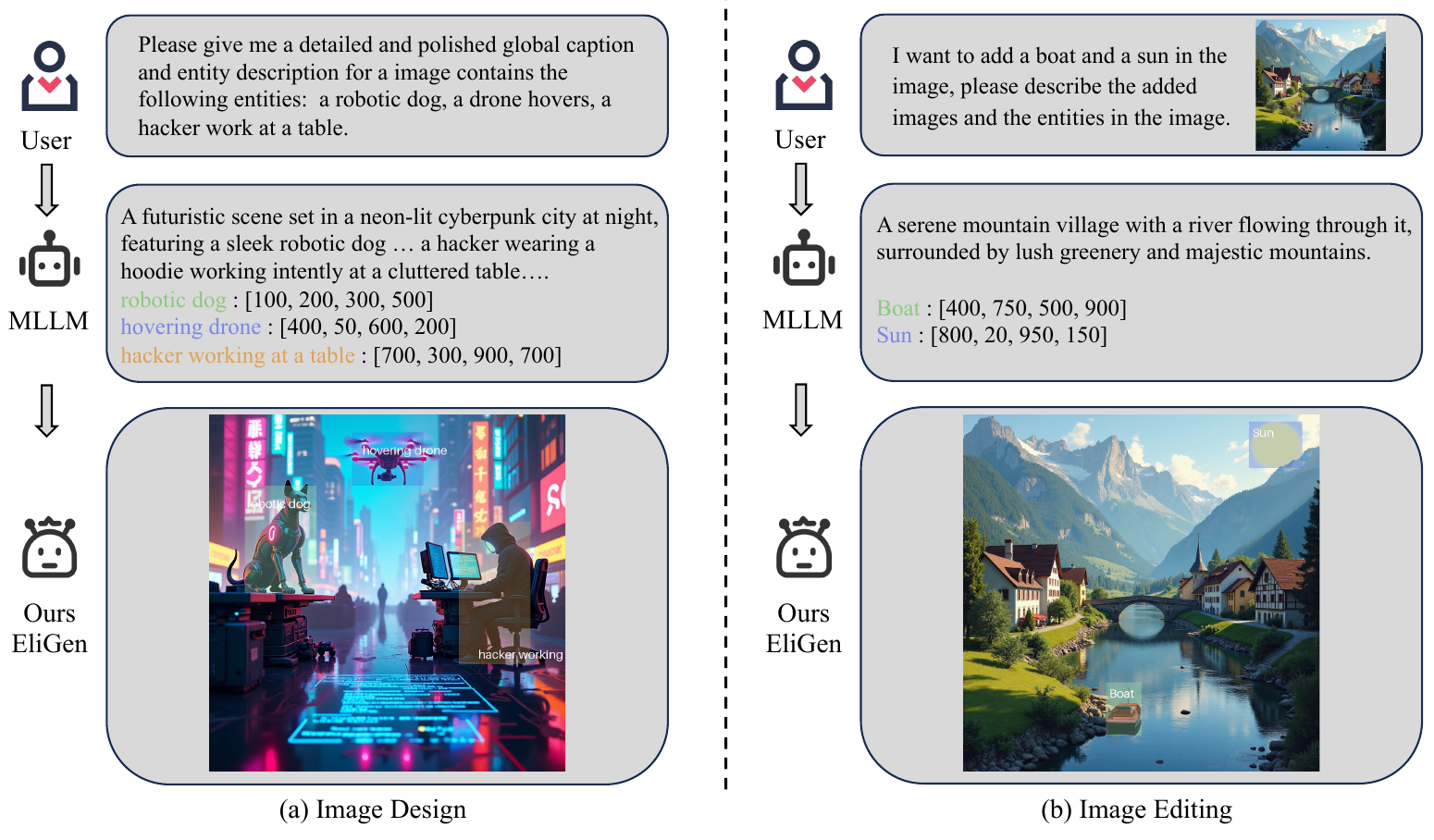}
    \caption{Dialogue-based image design and editing with MLLM and our EliGen. Users can utilize MLLM to design the layout of the image that they wish to generate or input the image that they want to edit into MLLM. After MLLM generates the caption and entity information of the image, EliGen can produce the target image.}
    \label{fig:mllm}
\end{figure*}

\section{Potential Creative Extension}
\label{Potential}
EliGen holds boundless creative potential, and here we demonstrate its capabilities when integrated with IP-Adapter \cite{flux-ipa}, In-Context LoRA \cite{lhhuang2024iclora} and MLLM \cite{wang2024qwen2vl}.

\begin{figure*}[thp] 
    \centering
    \includegraphics[width=1.\linewidth]{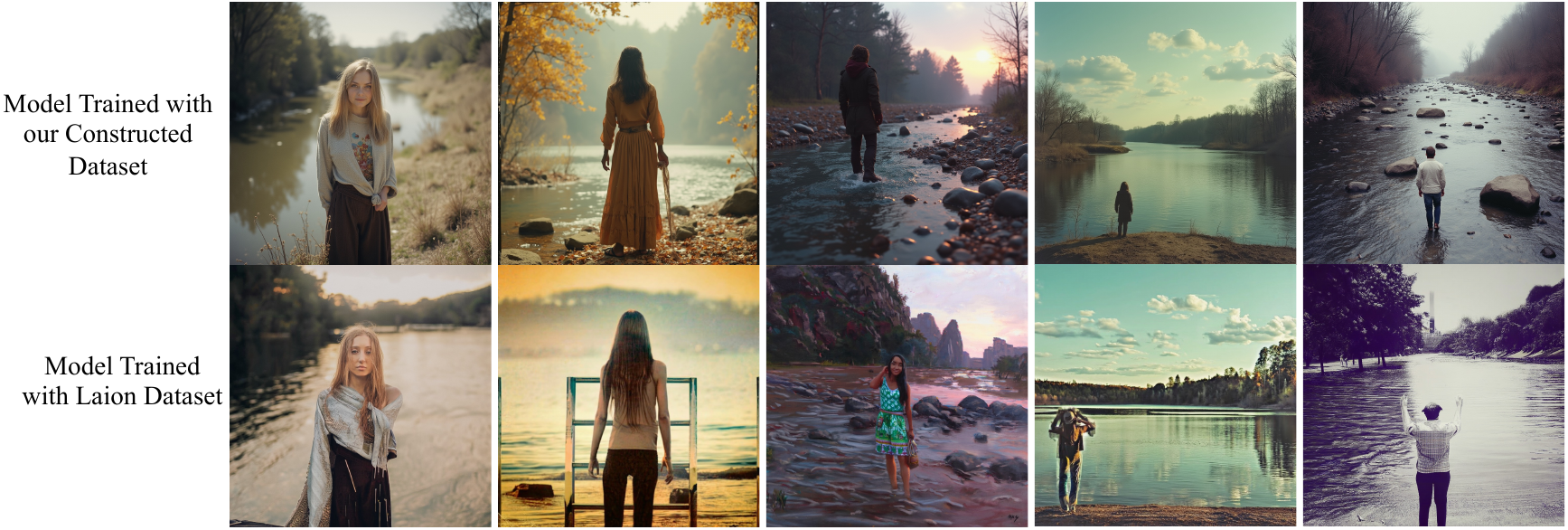}
    \caption{Training qualitative comparison between our constructed dataset and Laion dataset. The image pair is generated from the same noise, and the global prompt is ``A person standing by the river". The comparative results show that the images generated by the model trained on our proposed dataset significantly outperform those from the Laion dataset in quality.}
    \label{fig:laion}
\end{figure*}

\subsection{Styled Entity Control with IP-Adapter}
IP-Adapter \cite{flux-ipa} enables the generation of target images based on the style of a reference image. When combined with EliGen, it allows us to accomplish the styled entity control task. Figure~\ref{fig:IP-Adapter} illustrates some examples along with their generated variants. By inputting multiple entity details alongside a reference image, we achieve precise control over both the characters and the environmental style of the target image.

\subsection{Entity Transfer with In-Context LoRA}
In-Context LoRA \cite{lhhuang2024iclora} is capable of performing few-shot image composition tasks conditioned on either global prompts or images, and can be applied in various scenarios such as portrait photography, product design, font design, and visual identity design. Our EliGen model further enhances these scenarios by incorporating entity-level control conditions, enabling more personalized image generation. Taking visual identity design as an example, we have successfully accomplished the entity transfer task by introducing EliGen, as illustrated in Figure~\ref{fig:iclora}. Given the entity defined by the source image, we can transfer it to any desired location in the target image through EliGen's positional control capability, thereby achieving the intended visual identity.

\subsection{Dialogue-based Image Design and Editing with MLLM}
The MLLM is capable of interpreting user-provided images and textual content to engage in dialogue. As a result, we can achieve dialogue-based image design and editing by leveraging the integration of MLLM and EliGen. Specifically, grounding-enabled MLLMs such as Qwen2-VL \cite{wang2024qwen2vl} can be utilized to determine entity locations based on textual descriptions, enabling the generation of images solely through text using EliGen. Furthermore, the MLLM can be employed to comprehend images and design entity information in accordance with specific image editing requirements, thereby facilitating precise image editing in conjunction with EliGen. The qualitative results are shown in Figure~\ref{fig:mllm}.

\section{Ablation Study}

\subsection{Training Qualitative Comparison for Dataset Image Source}
The primary objective of training the EliGen model is to activate its capability for layout control, which necessitates the construction of a dataset annotated with entities. Moreover, the image distribution within the dataset should closely align with the output distribution of the model itself. If the distribution gap is substantial, a significant portion of the training loss would be attributed to this discrepancy. This would significantly impede the model's convergence and degrade the quality of the generated images. Consequently, we opt to utilize the Flux-generated images for dataset creation, rather than directly employing images from the open-source Laion dataset \cite{LAION-Aesthetics}.

We conduct training experiments using both our custom-created dataset and the Laion dataset, where the annotations for the Laion dataset are derived from the same pipeline described in Section~\ref{sec:pipeline}. As illustrated in Figure~\ref{fig:laion}, under identical prompts and initial noise conditions, the model trained on our dataset demonstrates significantly superior image quality compared to the model trained on the Laion dataset. Compared to Flux-generated images, the latter exhibits significant visual clutter, inferior detail quality, and a notable deviation from the expected distribution.

\begin{figure*}[thp] 
    \centering
    \includegraphics[width=1.\linewidth]{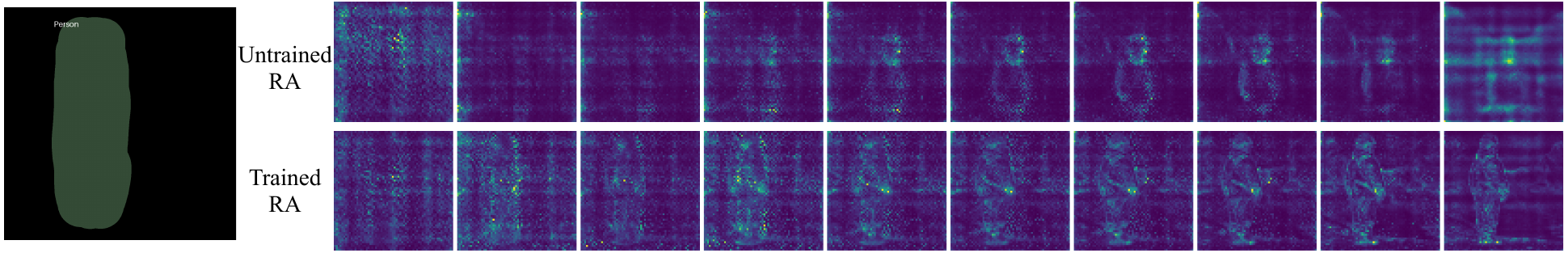}
    \caption{Attention maps of the local prompt ``person" with latent. The visualization results are derived from the regional attention (RA) layer at the last double-stream transformer block, showing the first ten denoising timesteps. As denoising progresses, the attention maps gradually reveal the shape of the entity, and the activation regions of the trained RA are primarily concentrated within the target control area.}
    \label{fig:attnmap}
\end{figure*}

\begin{figure*}[thp] 
    \centering
    \includegraphics[width=1.\linewidth]{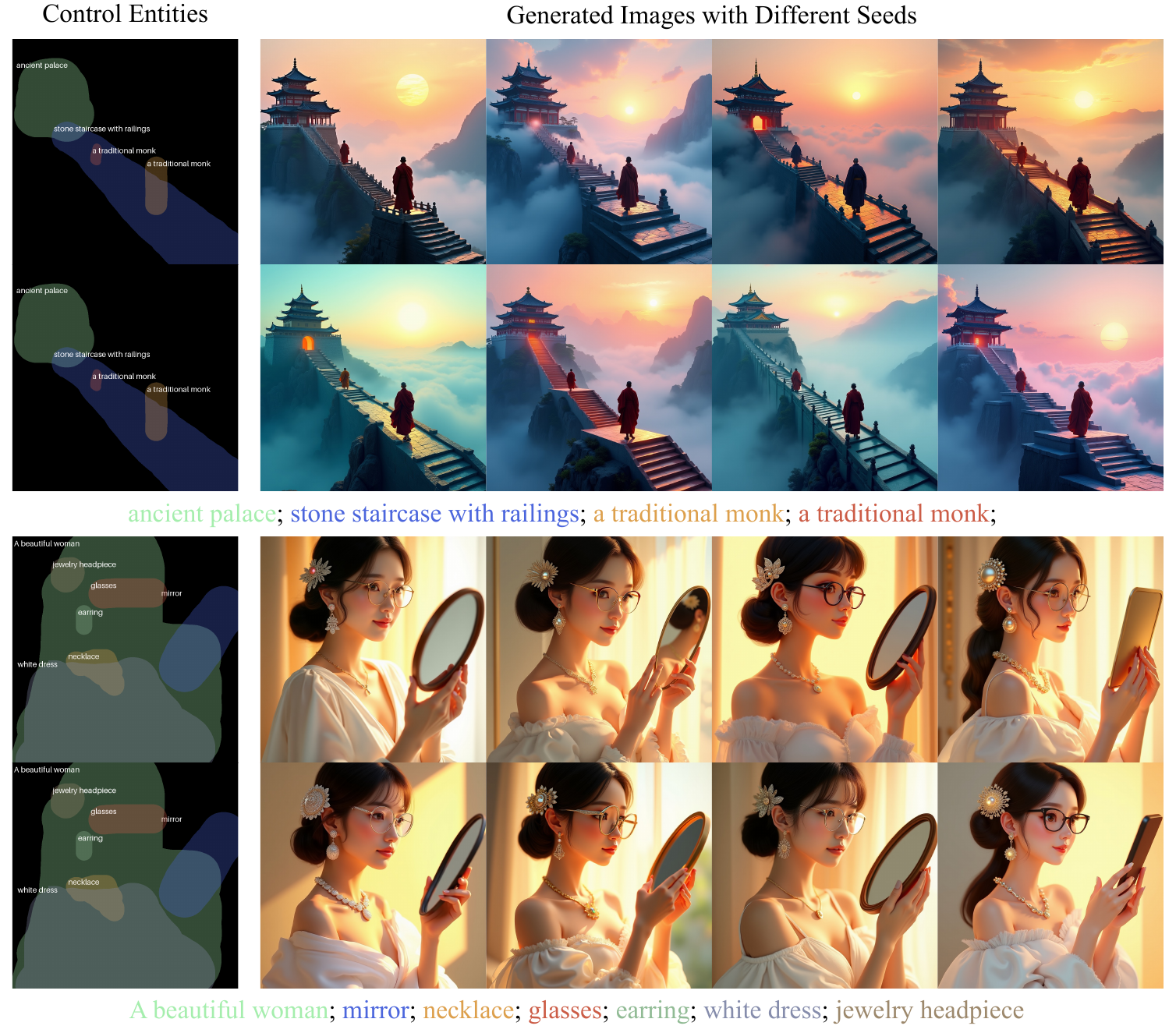}
    \caption{Generalization tests across different random seeds. Variations in the random seed induce changes in the initial noise input to the model. The two examples above demonstrate that EliGen possesses the capability to rectify images to the target layout for arbitrary noise inputs, while simultaneously preserving sufficient image diversity.}
    \label{fig:seed}
\end{figure*}

\subsection{Analysis on Attention Map}
Regional Attention (RA) primarily guides the generation of entities in specified regions through attention maps. Therefore, in Figure~\ref{fig:attnmap}, we visualize the attention maps to demonstrate that RA indeed plays a role in guiding entity generation. Note that here we visualize the attention maps of the entire latent image. In practice, during regional attention, the uncontrolled regions of the attention map are deactivated.

From the visualization results, it can be observed that the trained RA, compared to the untrained one, confines attention activation regions to the controlled area more rapidly. Meanwhile, as denoising progresses, the entity's shape gradually emerges within the controlled region, while activation in uncontrolled regions diminishes, thereby completing the controlled entity generation.

\section{Additional Qualitative Results}
EliGen is a robust entity-level control model, exhibiting strong generalization capabilities across diverse positional inputs, semantic contexts, and random noise. In this section, we present additional qualitative results to further validate its robustness.
\subsection{Various Random Seeds}
Variations in random seeds lead to changes in the initial noise input to the model. For conventional text-to-image models, the layout of the image is typically determined by the initial noise. In contrast, EliGen possesses the capability to control the layout, ensuring that each controlled entity is placed in the target position even when the initial noise varies. Test examples are shown in Figure~\ref{fig:seed}, where each image is generated with different random seeds, yet all adhere to the target layout, demonstrating the generalization ability of EliGen. Notably, the generated samples also exhibit diversity, as variations in random seeds result in changes in the details of each entity.
\subsection{Continuously Varying Entity Positions}
\begin{figure*}[thp] 
    \centering
    \includegraphics[width=1.\linewidth]{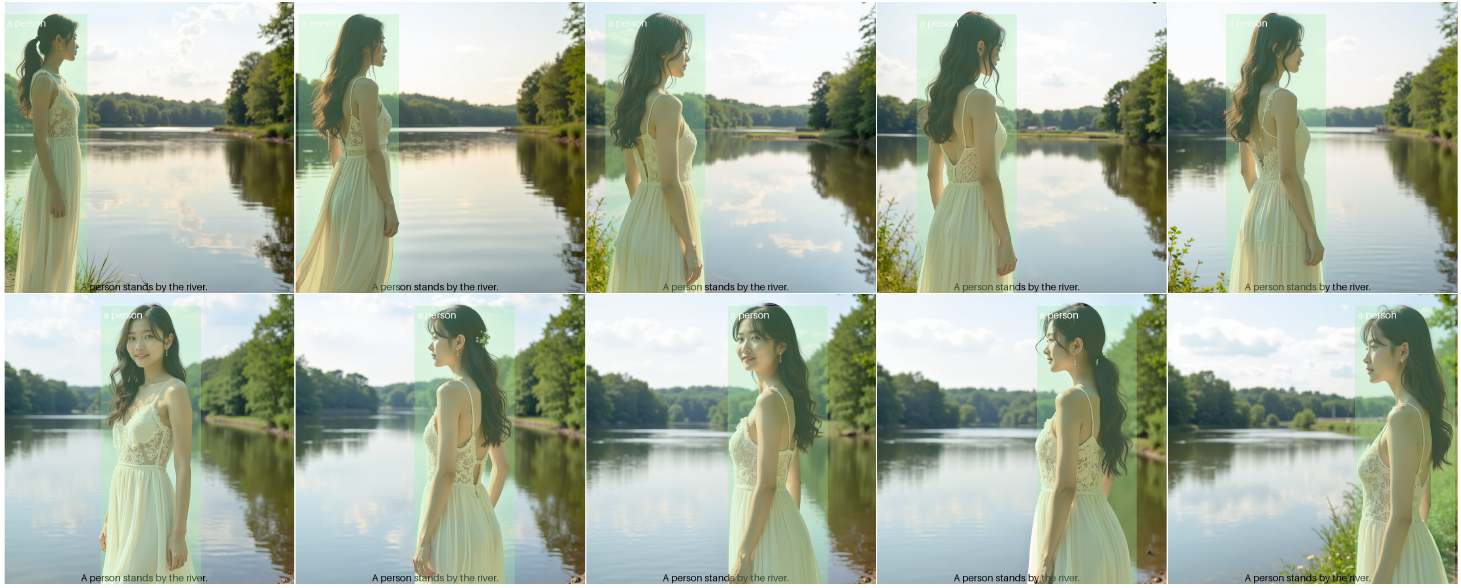}
    \caption{Continuous variation tests on positional inputs. The samples are generated using the same prompt and initial noise. EliGen can achieve continuous movement of entities, which highlights the potential of EliGen in generating coherent image sequences. To preserve the identity of the ``person" during generation, we integrate EliGen with the IP-Adapter method.}
    \label{fig:video}
\end{figure*}

In Figure~\ref{fig:video}, we evaluate the control efficacy of EliGen using identical input noise and the same entity. The position of the ``person" entity shifts progressively from left to right, with EliGen maintaining consistent control throughout the process. To preserve the identity of the person during generation, we integrate the IP-Adapter \cite{flux-ipa} method. This demonstrates EliGen's potential to generate coherent image sequences, such as those involving continuous movement of entities through positional control or complex actions orchestrated via coordinated control of multiple entities.

\subsection{Unreasonable Positional Inputs}
\begin{figure*}[thp] 
    \centering
    \includegraphics[width=1.\linewidth]{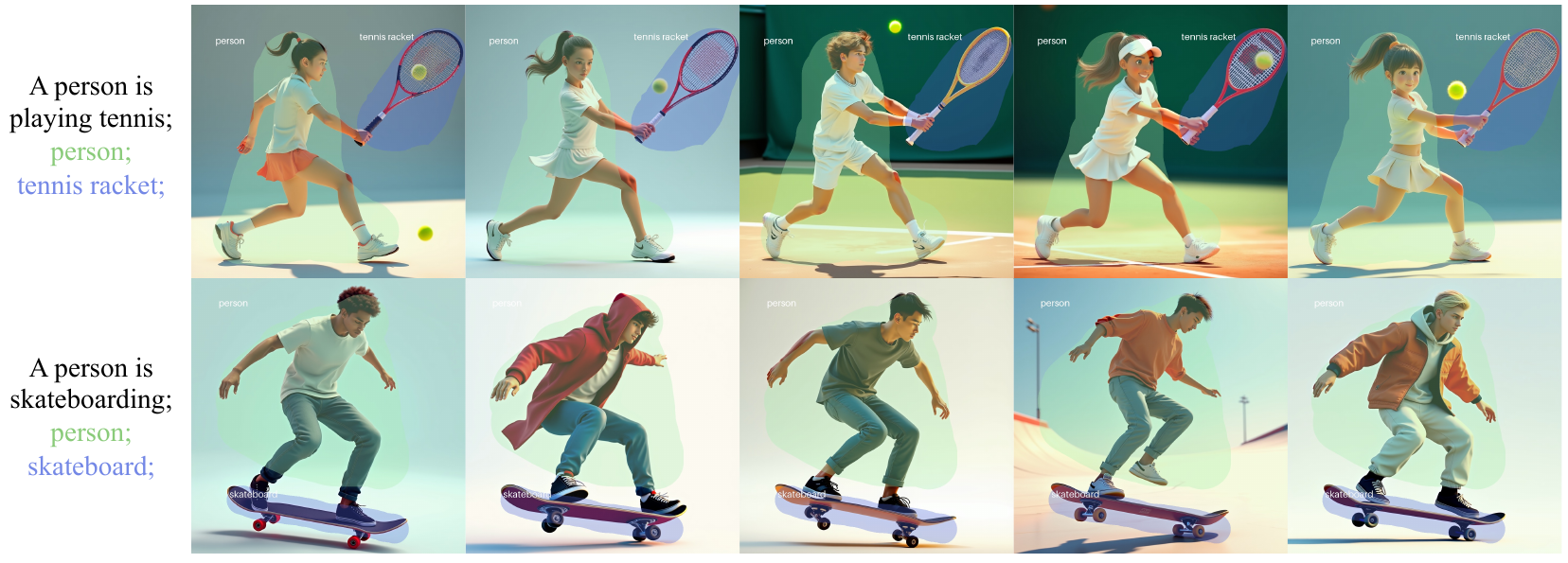}
    \caption{Generalization tests with incorrect spatial relationships between entities. In the provided examples, the input positional relationships between entities are inaccurate. For instance, the entities ``person" and ``tennis racket" should intersect rather than being separated to reflect the relationship of a person holding a tennis racket. Under such cases, EliGen prioritizes ensuring the coherence of entities in the image, thereby adaptively fine-tuning the actual positions of the entities, such as adjusting the position of the person's hand.}
    \label{fig:A4.3.1}
\end{figure*}

A robust model must demonstrate generalization capabilities even for suboptimal inputs. For EliGen, these suboptimal inputs primarily pertain to entity positions, which we categorize into two types: incorrect spatial relationships between entities and incorrect entity shapes.
\paragraph{Incorrect Spatial Relationships Between Entities}
incorrect spatial relationships between entities refer to scenarios where the relative positions of entities are inaccurate, such as being too far apart or too close together. We evaluate two cases of this inputs, as depicted in Figure~\ref{fig:A4.3.1}. In both examples, the correct positioning of the ``person" entity necessitates an overlap with another entity to accurately convey the action being performed. For example, in the prompt ``A person is playing tennis", the ``person" should ideally be holding a ``tennis racket". However, the input provided to EliGen is incorrect, which is featured two disjointed positions. In such scenarios, EliGen prioritizes the coherence of entities within the image, adaptively fine-tuning their actual positions, such as adjusting the position of the person's hand. Though this adaption may result in the model underperforming on quantitative metrics related to spatial accuracy, it substantially improves the overall image quality.

\begin{figure*}[thp] 
    \centering
    \includegraphics[width=1.\linewidth]{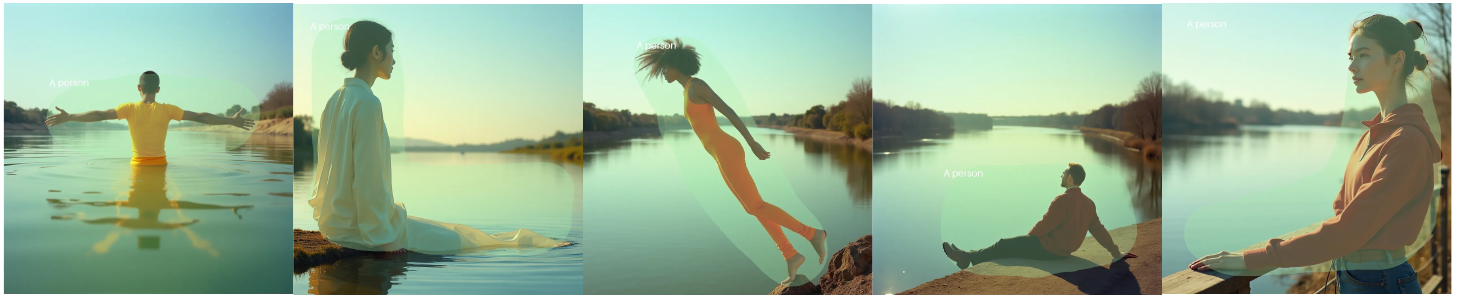}
    \caption{Generalization with incorrect entity shapes. EliGen demonstrates the ability to adaptively adjust entity morphology based on input shapes. In the examples, the control shape for the ``person" is not a standard upright stance. EliGen adjusts the person's posture according to the input mask to generate coherent entities.}
    \label{fig:A4.3.2}
\end{figure*}

\paragraph{Incorrect Entity Shapes}
Incorrect entity shapes refer to cases where the input region of an entity does not match its described shape, such as a ``standing person" typically being represented by a vertical rectangle shape. Through experimentation, we find that EliGen possesses the generalization to adaptively adjust entity posture based on arbitrary input shape. For example, in Figure~\ref{fig:A4.3.2}, EliGen adjusts the gesture of the generated ``person" entity according to various positional inputs while maintaining coherence with the environment.

\subsection{Pure Text-to-Image Generation Capability}
\begin{figure*}[thp] 
    \centering
    \includegraphics[width=1.\linewidth]{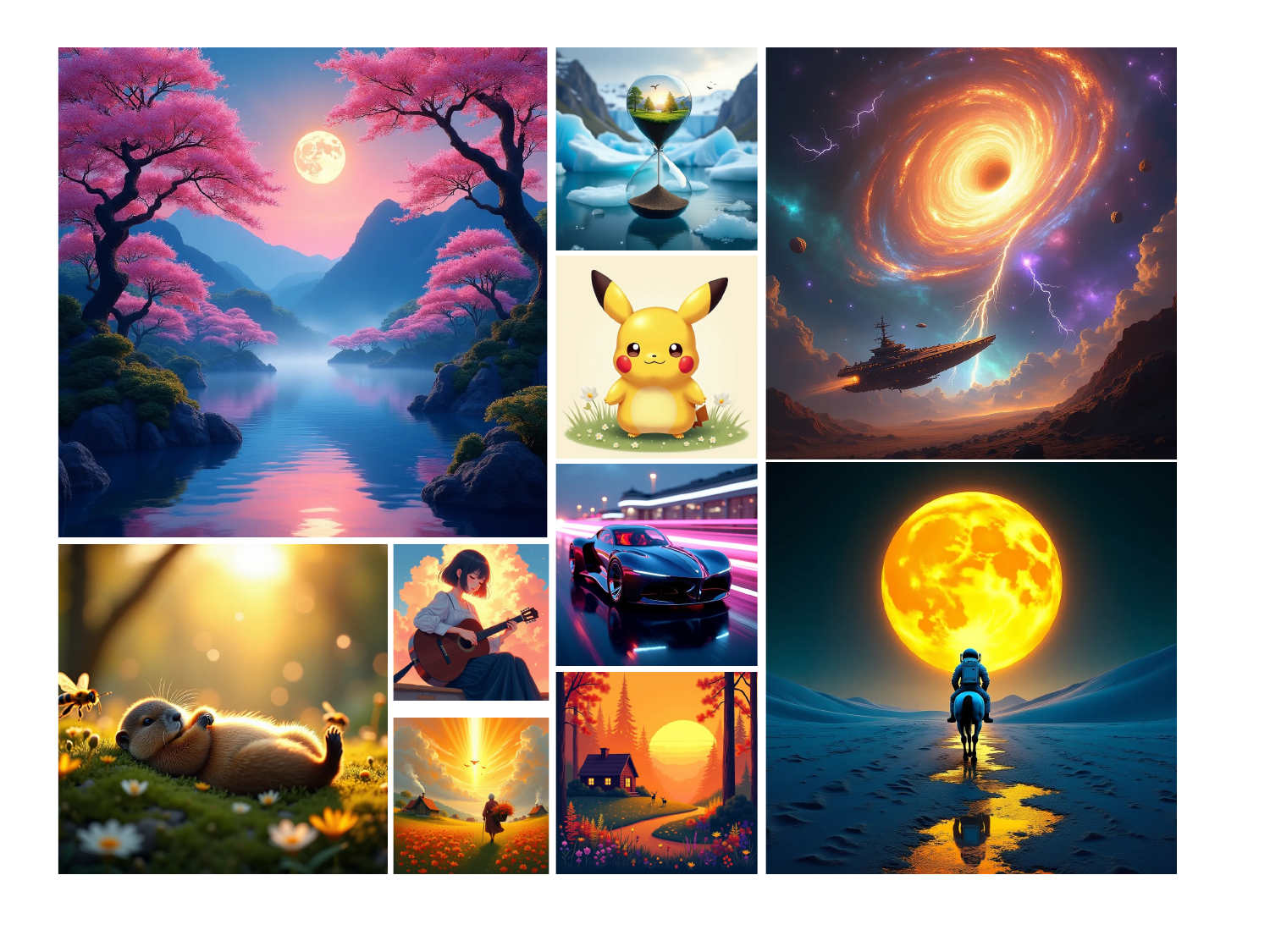}
    \caption{Pure text-to-image generation capability of EliGen. Despite fine-tuning on an entity-level control task, EliGen retains strong text-to-image generation performance without entity conditions, demonstrating that training does not degrade the model's image generation quality.}
    \label{fig:t2i}
\end{figure*}

The images generated by EliGen exhibit high quality. When provided with any number of entity inputs, it can achieve precise positional control and detailed generation for each entity. In the absence of any entity inputs, EliGen also functions as an exceptional text-to-image model. As shown in Figure~\ref{fig:t2i}, the images generated solely from global prompts demonstrate excellence in lighting, detail, and overall quality.


\end{document}